\documentclass[conference]{IEEEtran}
\IEEEoverridecommandlockouts

\usepackage{cite}
\usepackage{amsmath,amssymb,amsfonts}
\usepackage{algorithmic}
\usepackage{graphicx}
\usepackage{textcomp}
\usepackage{xcolor}
\usepackage[utf8]{inputenc}
\usepackage[T1]{fontenc}
\usepackage{url}

\usepackage{multirow}
\usepackage{amssymb}
\newcommand{\cmark}{\checkmark}
\newcommand{\xmark}{$\times$}

\def\BibTeX{{\rm B\kern-.05em{\sc i\kern-.025em b}\kern-.08em
    T\kern-.1667em\lower.7ex\hbox{E}\kern-.125emX}}
\begin{document}

\title{Privacy Evaluation of Generative Models for Trajectory Generation\\

\thanks{This work was supported by the MUSIT Project through the European Union’s Horizon Europe Framework Programme (HORIZON), under Marie Sklodowska-Curie grant agreement no. 101182585. The work only reflects the authors’ views; the EU Agency is not responsible for any use of the information it contains.}
}

\author{

\IEEEauthorblockN{
Stavros Bouras$^{1}$,
Ioannis Kontopoulos$^{1}$,
Chiara Pugliese$^{3}$,
Francesco Lettich$^{2}$,\\
Emanuele Carlini$^{2}$,
Hanna Kavalionak$^{2}$,
Chiara Renso$^{2}$,
Konstantinos Tserpes$^{1}$
}

\IEEEauthorblockA{\textit{$^{1}$School of Electrical and Computer Engineering, National Technical University of Athens, Athens, Greece}}

\IEEEauthorblockA{\textit{$^{2}$Institute of Information Science and Technologies, National Research Council (CNR), Pisa, Italy}}

\IEEEauthorblockA{\textit{$^3$Institute of Informatics and Teletematics, National Research Council (CNR), Pisa, Italy}}

}

\maketitle

\begin{abstract}

Trajectory data is fundamental to modern urban intelligence, yet its sensitivity raises significant privacy concerns. Generative models such as Generative Adversarial Networks, Variational Autoencoders, and Diffusion Models have been developed to generate realistic synthetic trajectory data by capturing underlying spatiotemporal distributions and mobility patterns. Although these models are often assumed to preserve privacy due to their generative nature, this assumption does not necessarily hold. In this work, we investigate the intersection of generative trajectory modeling and privacy evaluation. By identifying applicable empirical methods for assessing privacy preservation in trajectory generation tasks, we demonstrate a significant gap in the evaluation of privacy for generative trajectory models. Motivated by this gap, we implement Membership Inference  Attacks against representative models, demonstrating the feasibility of using such empirical privacy evaluation methods 
and showing that their generative nature does not eliminate privacy risks.
\end{abstract}


\section{Introduction}


The widespread availability of location-aware devices has made trajectory data a valuable resource for urban applications such as traffic management and route planning. At the same time, trajectory data is deeply sensitive, as few as four spatiotemporal points are sufficient to uniquely identify 95\% of individuals \cite{de2013unique}. To address this limitation, the use of deep generative models  has rapidly increased for synthetic trajectory generation \cite{chen2025trajectory}. The use of deep learning models helps capture complex spatiotemporal dependencies and allows us to avoid sharing real data, preserving user privacy. However, this assumption is not guaranteed, as generative models are known to memorize aspects of their training data \cite{carlini2019secret}, leading their outputs to resemble, or even reconstruct, the original training records, making individuals vulnerable to inference attacks, demonstrated in works such as \cite{buchholz2024sok}. This highlights the need for empirical privacy evaluation of generative trajectory models, enabling a comprehensive assessment from multiple perspectives and ensuring that the generated synthetic data does not expose the privacy of the individuals from whom it was derived. 

In this paper, we aim to provide a systematic investigation of privacy in synthetic trajectory generation, with a particular focus on how privacy risks can be assessed in practice by introducing the necessary background at the intersection of trajectory generation and privacy, covering fundamental concepts, threat models, and formal privacy guarantees.

Our contributions are threefold. First, we identify and categorize 
the empirical privacy evaluation methods applicable to the trajectory 
generation domain (Section~\ref{subsec:methods}). Second, we examine 
representative generative models, spanning variational autoencoders, 
generative adversarial networks, and diffusion models, and 
systematically map them to the identified evaluation methods, 
highlighting a significant gap in current practices 
(Section~\ref{sec:privacy_eval}). Third, we demonstrate the 
feasibility of membership inference attacks against generative 
trajectory models, showing that privacy risks are tangible and 
that empirical evaluation is both necessary and practical 
(Section~\ref{sec:mia}). 

    
    
    

\section{Preliminaries}

We introduce the concept of trajectory and its generation below.

\subsubsection{Trajectory Definition}
Formally, a trajectory is defined as  a series of spatiotemporal points $S = \{x_1, x_2, \ldots, x_n\} \in \mathbb{R}^{N \times 2}$, 
where each element is represented as a tuple $(l_i, t_i)$, in which 
$l_i$ denotes the spatial location and $t_i$ denotes the timestamp 
of the $i$-th  \cite{kong2023mobility}. 

\subsubsection{Trajectory Generation}
Trajectory generation is defined as the process where, given a trajectory dataset $\mathcal{D} = \{\tau_1, \tau_2, \ldots, \tau_N\}$ and a set of environmental constraints $\mathcal{C}$, the goal is to synthesize a new set of trajectories $\mathcal{D}' = \{\tau_1', \tau_2', \ldots, \tau_N'\}$ such that the generated set preserves the spatiotemporal properties of the original dataset, i.e., the distributions $P(\mathcal{D})$ and $P(\mathcal{D}')$ are similar \cite{chen2025trajectory}.

The three main approaches used to generate synthetic trajectories are Variational Autoencoders (VAEs), Generative Adversarial Networks (GANs), and Diffusion Models. VAEs \cite{kingma2013auto} use an encoder-decoder architecture, mapping input data to a low-dimensional latent space from which new samples are generated. GANs \cite{goodfellow2020generative} employ two competing networks - a generator that produces synthetic samples and a discriminator that tries to distinguish them from real ones - learning the data distribution through this adversarial process. Diffusion Models \cite{sohl2015deep, ho2020denoising, song2020score, song2020denoising} are iterative generative models that first corrupt data by progressively adding noise, and then learn to recover the original data through a gradual reverse denoising process.

\subsection{\textbf{Privacy}}
Machine learning models tend to memorize information during training, which can expose sensitive data during generation and pose privacy risks. \cite{carlini2019secret} distinguishes three related but distinct concepts: \textit{overfitting}, which occurs when a model learns training data too well and fails to generalize; \textit{overtraining}, which refers to the specific point at which validation error stops decreasing; and \textit{unintentional memorization}, defined as the retention of out-of-distribution training data that is irrelevant to the learning task and independent of overtraining, beginning early in the training process and peaking when validation loss is minimized. These distinctions highlight that some degree of memorization is an inherent and unavoidable part of training, occurring even in well-trained models that show no signs of overfitting or overtraining. This behavior extends directly to generative models, where 
memorization during training can surface during generation 
and expose sensitive information in synthetic data applications.

When the data synthetically generated are trajectories, significant privacy risks emerge \cite{jin2022survey} because trajectory data differs fundamentally from other modalities. It is constrained by spatial structures and mobility patterns, exhibits strong spatiotemporal continuity, and carries semantic meaning through points of interest that can reveal sensitive personal attributes. Critically, anonymized trajectories can be re-identified from as few as 4 spatiotemporal points, sufficient to uniquely identify 95\% of individuals \cite{de2013unique}. Furthermore, despite generative models producing synthetic samples, the generated data cannot be assumed to be private \cite{miranda2024overview}, as privacy can be compromised when generated samples closely resemble training data or when phenomena such as memorization occur during training, enabling adversaries to infer sensitive information through privacy attacks.

\subsubsection{\textbf{Privacy Attacks}}
\label{subsubsec:attacks}

In privacy attacks, adversaries seek to infer information that was not intended to be disclosed \cite{rigaki2023survey}. This information can range from details about the training data of the targeted machine learning model to properties of the data or information about the model itself. As presented in \cite{rigaki2023survey}, a taxonomy of privacy attacks includes the following types of attacks: 

\begin{itemize}
    \item \textit{Membership Inference Attacks (MIAs)} \cite{shokri2017membership}, which aim to determine if a specific data sample was part of the original training dataset of a machine learning model. 
    \item \textit{Reconstruction Attacks}, also referred to as \textit{Model Inversion Attacks} \cite{fredrikson2014privacy,fredrikson2015model}, that seek to recreate training samples either fully or partially, with the possibility of also recovering their corresponding training labels.
    \item \textit{Property Inference Attacks} \cite{ateniese2015hacking}, which aim to infer dataset-level properties that are neither explicitly represented as features nor directly tied to the learning task. These properties (e.g., demographic characteristics) can be exposed, as models unintentionally learn such information, even in well-generalized models, since it is an inherent byproduct of the learning process \cite{carlini2019secret,rigaki2023survey}.
    \item \textit{Model Extraction Attacks} \cite{tramer2016stealing} that seek to reconstruct a substitute model that closely replicates the behavior of the target model, either by matching its task accuracy or approximating its decision boundary.
\end{itemize}

While the above taxonomy characterizes the broad machine learning field, the trajectory-data domain introduces additional domain-specific privacy threats due to its spatiotemporal properties. In particular, in \cite{jin2022survey}, trajectory-specific privacy attack types are identified, distinguishing between \textit{Linkage Attack Models} and \textit{Probabilistic Attack Models}. The former focuses on the type of sensitive data being inferred, while the latter focuses on how much information is revealed. \textit{Linkage Attacks} can be further divided into \textit{Record Linkage}, which seeks to infer individual identity (e.g., re-identification), \textit{Attribute Linkage}, which aims to infer personal profile information by leveraging frequent occurrences among similar trajectories, \textit{Table Linkage}, which determines whether a specific individual is present in a dataset (a threat closely related to membership inference attacks), and \textit{Group Linkage}, which focuses on extracting social relationships between individuals from their trajectory data.

\subsubsection{\textbf{Formal Privacy Guarantees - Differential Privacy}}
\label{subsubsec:dp}

Existing defenses against privacy attacks do not provide formal guarantees and are limited to specific attack scenarios rather than offering comprehensive protection.

\textit{Differential Privacy (DP)} \cite{dwork2006calibrating, dwork2014algorithmic} is the only framework that provides such guarantees, ensuring that the contribution of any individual record to the output of a learning algorithm is formally bounded. Formally, DP is defined such that:

A randomized mechanism $\mathcal{M}: \mathcal{D} \rightarrow \mathcal{S}$ is $(\varepsilon, \delta)$-differentially private if for any two adjacent datasets $D, D' \in \mathcal{D}$ differing by at most one record, and for any subset of outputs $S \subseteq \mathcal{S}$:
\begin{equation}
    \Pr[\mathcal{M}(D) \in S] \leq e^{\varepsilon} \cdot \Pr[\mathcal{M}(D') \in S] + \delta
\end{equation}
where $\varepsilon > 0$ is the privacy budget controlling the strength of the privacy guarantee, and $\delta \geq 0$ is the failure probability. The definition of adjacent datasets is central to the DP framework, as it determines what constitutes the record whose addition or removal must not significantly change the output of the mechanism. This record, therefore, is the unit of data that receives the privacy protection of the framework. 

In the context of Deep Learning, DP is implemented through \textit{Differentially Private Stochastic Gradient Descent (DP-SGD)} \cite{abadi2016deep, ponomareva2023dp}, which incorporates the DP framework in the training process at the level of the training algorithm. DP-SGD operates by first bounding the influence of individual training samples through gradient clipping, then adding calibrated Gaussian noise to the clipped gradients, while a privacy accountant tracks the cumulative privacy budget $\varepsilon$ to ensure the desired privacy guarantee.

In the \textit{trajectory domain}, this unit of data has been referred to as the \textit{Unit of Privacy (UoP)} \cite{buchholz2024sok}, and it should be carefully chosen as it introduces a trade-off: a larger UoP requires more obfuscation to achieve the same level of privacy, thereby increasing utility loss, while a smaller UoP implies protecting a smaller unit of information, leading to increased risks of correlation attacks \cite{miranda2023sok} or reconstruction attacks \cite{buchholz2022reconstruction}. In \cite{buchholz2024sok}, four levels of UoP are identified, each representing a different definition of neighboring datasets in the DP mechanism: 
\begin{itemize}
    \item \textit{User-Level}, where $D'$ differs from $D$ by the removal of all trajectories associated with a specific user, providing the highest level of theoretical protection but at the cost of a significant utility trade-off.
    
    \item \textit{Instance-Level}, also referred to as \textit{trajectory-level}, where $D'$ differs by one complete trajectory, thus protecting the trajectory as a unit and hindering the exploitation of intra-trajectory correlations that many attacks rely on. This is the most common level of privacy in deep learning, as DP-SGD provides this level of UoP for training samples.
    
    \item \textit{Location-Level}, where the unit of protection is an individual location within a trajectory, offering the weakest level of privacy. In this setting, works on DP-based trajectory publication \cite{jiang2013publishing} and synthetic generation \cite{kim2022deep} have been proposed. However, the level of protection is not guaranteed to extend to the full trajectory, as independently protecting each location is only effective when the number of trajectory points is small and does not scale well to longer trajectories \cite{andres2013geo}. It is therefore suggested that better privacy is achievable by protecting the trajectory as a whole rather than its individual locations \cite{andres2013geo}.

    \item \textit{Multi-Event-Level}, which lies between instance- and location-level privacy by protecting a window of multiple locations within a trajectory.
\end{itemize}

Ultimately, the choice of UoP and, more generally, the incorporation of DP in synthetic data generation introduces the privacy–utility trade-off, as stronger privacy guarantees degrade the utility of the generated data \cite{rigaki2023survey}. Specifically, in the trajectory domain, this is particularly evident, as the spatiotemporal structure of the data makes it sensitive to noise addition \cite{buchholz2024sok}.

Beyond the privacy-utility trade-off, DP cannot defend against every privacy attack and its protection is bounded in scope. While it provides formal guarantees against Membership Inference and Reconstruction attacks \cite{rigaki2023survey}, it cannot provide such protection against property inference attacks \cite{ateniese2015hacking}. Furthermore, model extraction attacks focus on replicating the model's functionality rather than extracting information from the training data, and therefore defenses rely more on query detection and prediction obfuscation than on the data-level protection provided by DP. 

Moreover, even in cases where DP is applied, as demonstrated in the trajectory domain, some implementations contain flawed DP proofs, meaning that the claimed guarantees do not hold, despite the underlying framework being sound \cite{buchholz2024sok}. Finally, even correct implementations of DP can be undermined by an inappropriate choice of UoP \cite{buchholz2024sok}.

These boundaries of formal privacy guarantees indicate that no single defensive mechanism can provide comprehensive protection against all privacy-related threats. Privacy attacks that fall outside the scope of DP represent an attack surface that lacks formal guarantees and therefore require empirical evaluation, motivating their use as tools for assessing privacy.

\subsection{\textbf{Empirical Privacy Evaluation of Trajectories}}
\label{subsec:methods}

To address these limitations, a variety of empirical evaluation 
approaches have been developed for the trajectory domain, ranging 
from adversarial attacks that directly exploit model vulnerabilities 
to quantitative metrics that measure the degree of privacy 
preservation. These include Trajectory User Linking (TUL) 
\cite{gao2017identifying}, Linkage Attack Probability (LA) 
\cite{jin2020trajectory}, Reconstruction Attacks such as RAoPT 
\cite{buchholz2022reconstruction} and iTracker 
\cite{shao2020structured}, Tracking Attack Probability (TA) 
\cite{shokri2011quantifying, zhang2025dp}, Mutual Information 
(MI) \cite{cover1999elements} and Membership Inference Attacks 
(MIA). Among these, MIAs are of particular interest as they 
directly assess whether a model has memorized specific training 
samples, representing one of the most critical privacy risks in 
machine learning \cite{carlini2019secret}. The application of 
MIAs to generative trajectory models is discussed in detail in 
Section~\ref{sec:mia}.

\section{Privacy Evaluation in Generative Trajectory Models}
\label{sec:privacy_eval}

To contextualize the privacy evaluation gap in the trajectory 
generation literature, representative generative models spanning 
VAEs, GANs and Diffusion Models were examined with respect to 
the privacy evaluation methods identified in 
Section~\ref{subsec:methods}. VAE-based and Diffusion Model-based 
works were found to include no privacy evaluation, focusing 
instead on generation quality and utility. This is consistent 
with the common view that synthetic data generation is inherently 
privacy-preserving, a claim that as discussed in 
Section~\ref{subsec:methods}, is not always guaranteed. Among 
GAN-based models, only a small subset includes any form of 
privacy evaluation, as shown in Table~\ref{tab:privacy_eval}. 
Notably, the MIA column remains entirely empty across all models, 
confirming the gap that this work addresses.

\begin{table}[h]
\centering
\caption{Privacy Evaluation Methods Used by Generative Trajectory Models}
\label{tab:privacy_eval}
\resizebox{\columnwidth}{!}{%
\begin{tabular}{llccccccc}
\hline
\textbf{Family} & \textbf{Model} & \textbf{TUL} & \textbf{LA} & \textbf{RAoPT} & \textbf{iTracker} & \textbf{TA} & \textbf{MI} & \textbf{MIA} \\
\hline
VAEs & \multicolumn{8}{c}{$-$} \\
\hline
\multirow{4}{*}{GANs}
& LSTM-TrajGAN \cite{rao2020lstm} & \cmark & \xmark & \cmark$^{\dagger}$ & \xmark & \xmark & \xmark & \xmark \\
& LGAN-DP \cite{zhang2023lgan} & \xmark & \xmark & \xmark & \xmark & \xmark & \cmark & \xmark \\
& DP-TrajGAN \cite{zhang2023dp} & \xmark & \xmark & \xmark & \xmark & \xmark & \cmark & \xmark \\
& DP-LTGAN \cite{zhang2025dp} & \xmark & \cmark & \xmark & \xmark & \cmark & \xmark & \xmark \\
\hline
DMs & \multicolumn{8}{c}{$-$} \\
\hline
\multicolumn{9}{l}{\footnotesize \cmark: evaluated, \xmark: not evaluated, $-$: no privacy evaluation conducted.} \\
\multicolumn{9}{l}{\footnotesize $^{\dagger}$RAoPT evaluation for LSTM-TrajGAN conducted in \cite{buchholz2024sok}.} \\
\hline
\end{tabular}%
}
\end{table}

Among the identified evaluation methods in Section~\ref{subsec:methods}, MIAs are particularly suitable for evaluating the privacy of generative trajectory models, as they directly probe whether specific training samples have been memorized by the model. Such attacks can exploit the inherent spatiotemporal structure of trajectory data, demonstrating that generative models are not inherently immune to inference attacks. This is further reinforced by the limitations of formal privacy guarantees discussed in Section~\ref{subsubsec:dp}, where flawed DP proofs and inappropriate Unit of Privacy selections leave models vulnerable to precisely such inference threats, even when a formal guarantee is claimed. Together, these observations suggest that empirical privacy evaluation approaches should be applied more broadly and systematically to generative trajectory models, covering the full range of methods outlined in Section~\ref{subsec:methods} in order to properly assess their privacy.  Motivated by this, the following section showcases the feasibility of MIAs against generative trajectory models as a first step toward addressing this gap.

\section{Empirical Experimental Privacy Evaluation via Membership Inference Attacks}
\label{sec:mia}

\subsection{Threat Model}

As discussed in Section~\ref{subsubsec:attacks}, the overall goal 
of MIAs is to determine, given a trained model $\mathcal{M}_\theta$ 
and a target sample $x$, whether $x \in D_{train}$ (member) or 
$x \notin D_{train}$ (non-member), where $D_{train}$ is the training 
dataset of $\mathcal{M}_\theta$. In this work, we apply MIAs against 
four representative generative trajectory models, namely two 
GAN-based models, LSTM-TrajGAN \cite{rao2020lstm} and MoveSim 
\cite{feng2020learning}, and two Diffusion Model-based models, 
DiffTraj \cite{zhu2023difftraj} and Diff-RNTraj \cite{wei2024diff}. 
A white-box setting is considered, in which the adversary has full 
access to the trained model's weights and architecture.

\subsection{Attack Formulations}

\subsubsection{Discriminator-based MIA on GAN-based Trajectory Models}

Based on \cite{chen2020gan}, which provides a systematic taxonomy of 
MIAs against generative models, we implement a discriminator-based 
MIA against LSTM-TrajGAN \cite{rao2020lstm} and MoveSim 
\cite{feng2020learning}. The discriminator $\mathcal{D}$, trained to 
distinguish real training trajectories from synthetic ones, serves as 
the membership signal. If $\mathcal{D}$ overfits to the training data, 
it will assign systematically higher confidence to members than to 
non-members, making its output a membership indicator 
\cite{chen2020gan}. For a target trajectory $x$, the membership score 
is computed as:

\begin{equation}
s(x) = \log\frac{\mathcal{D}(x)}{1 - \mathcal{D}(x)}
\end{equation}

where the logit transformation is applied to $\mathcal{D}$'s output 
to improve membership separation \cite{carlini2022membership}. Attack 
performance is evaluated using AUC-ROC \cite{chen2020gan}.

\subsubsection{Loss-based MIA on Diffusion-based Trajectory Models}

Grounded in the theoretical connection between overfitting and 
membership leakage \cite{yeom2018privacy}, we adopt a loss-based 
MIA against DiffTraj \cite{zhu2023difftraj} and Diff-RNTraj 
\cite{wei2024diff}, following the white-box loss-based attack 
formulation for diffusion models proposed in \cite{matsumoto2023membership} 
and \cite{hu2023membership}. If a model overfits to its training 
data, it will assign systematically lower denoising loss to members 
than to non-members. For a target trajectory $x_0$, the membership 
score is computed as the negative average noise prediction MSE over 
multiple randomly sampled diffusion timesteps:

\begin{equation}
s(x_0) = -\frac{1}{T}\sum_{t=1}^{T} \mathbb{E}_{\varepsilon} 
\left[\|\varepsilon - \varepsilon_\theta(x_t, t)\|^2\right]
\end{equation}

where $\varepsilon_\theta$ is the model's noise prediction network, 
$\varepsilon \sim \mathcal{N}(0, \mathbf{I})$ is the Gaussian noise 
added at timestep $t$, $x_t$ is the noisy trajectory at timestep 
$t$, and $T$ is the number of probe timesteps. Attack performance is evaluated using AUC-ROC.

\subsection{Datasets}

The GAN-based models are evaluated on the datasets used in their 
original training procedures. LSTM-TrajGAN \cite{rao2020lstm} was 
trained on the Foursquare NYC weekly trajectory dataset 
\cite{yang2014modeling, may2020marc}, while MoveSim 
\cite{feng2020learning} was trained on the GeoLife dataset 
\cite{zheng2010geolife}. Both diffusion-based models, DiffTraj 
\cite{zhu2023difftraj} and Diff-RNTraj \cite{wei2024diff}, were 
trained on the DiDi Chengdu taxi trajectory 
dataset\footnote{\url{https://outreach.didichuxing.com}}.

\subsection{Implementation Details}

For each model, membership scores are computed on a set of member 
trajectories, drawn from the training data, and non-member 
trajectories, drawn from a held-out set that was never seen by 
the model during training. For the diffusion-based models, 
membership scores are averaged over 50 randomly sampled timesteps 
from the full diffusion range, following \cite{hu2023membership}, 
to produce a stable and model-agnostic signal. The implementation 
details for each model are summarized in Table~\ref{tab:mia_setup}.

\begin{table}[h]
\centering
\caption{MIA Implementation Details}
\label{tab:mia_setup}
\begin{tabular}{llll}
\hline
\textbf{Model} & \textbf{Attack} & \textbf{Eval. Set} & \textbf{Timesteps} \\
\hline
LSTM-TrajGAN \cite{rao2020lstm} & Disc. score & 2,052 / 1,027 & -- \\
MoveSim \cite{feng2020learning} & Disc. score & 7,000 / 2,500 & -- \\
DiffTraj \cite{zhu2023difftraj} & Loss-based & 2,000 / 2,000 & 50 random \\
Diff-RNTraj \cite{wei2024diff} & Loss-based & 2,000 / 2,000 & 50 random \\
\hline
\multicolumn{4}{l}{\footnotesize Eval. Set: members / non-members.} \\
\hline
\end{tabular}
\end{table}

\subsection{Results}

AUC-ROC measures the probability that the attack assigns a higher 
membership score to a training sample than to a non-training sample, 
with a value of 0.5 indicating performance equivalent to random 
guessing and values above 0.5 indicating exploitable membership 
leakage.

\begin{table}[h]
\centering
\caption{MIA Results on Generative Trajectory Models}
\label{tab:mia_results}
\begin{tabular}{llc}
\hline
\textbf{Model} & \textbf{Family} & \textbf{AUC-ROC} \\
\hline
LSTM-TrajGAN \cite{rao2020lstm} & GAN & 0.5029 \\
MoveSim \cite{feng2020learning} & GAN & \textbf{0.7002} \\
DiffTraj \cite{zhu2023difftraj} & DM & 0.5012 \\
Diff-RNTraj \cite{wei2024diff} & DM & 0.4949 \\
\hline
\end{tabular}
\end{table}

As can be seen from Table~\ref{tab:mia_results}, the results 
vary across models. MoveSim exhibits an AUC-ROC of 0.70, 
indicating clear evidence of membership leakage under the 
implemented attack and suggesting that the model has memorized 
specific training trajectories to a degree that an adversary 
can exploit. LSTM-TrajGAN, DiffTraj, and Diff-RNTraj, by 
contrast, show AUC-ROC values close to 0.5, consistent with 
random guessing, indicating that the implemented attacks were 
unable to extract meaningful membership signals. This suggests 
that, under the evaluated attack formulations, no exploitable 
membership leakage was observed in these models.

These results demonstrate the feasibility of implementing MIAs as 
an empirical privacy evaluation tool for generative trajectory models. 
The finding that MoveSim exhibits membership leakage while the 
remaining models show resistance confirms that privacy risks in 
generative trajectory models are real and model-dependent, and cannot 
be assumed absent simply because the model produces synthetic data. 
The attacks implemented in this work represent a single class of 
empirical privacy evaluation methods. Other MIA formulations, as well 
as other categories of privacy attacks identified in 
Section~\ref{subsec:methods}, could reveal additional vulnerabilities 
not captured here, further underlining the need for comprehensive 
empirical privacy evaluation of generative trajectory models.

\section{Conclusions \& Future work}
This work examined the privacy of generative trajectory models 
from an empirical perspective, identifying a significant gap in 
how privacy is assessed in the trajectory generation literature. 
Although many of the examined models were not developed with 
privacy as a primary concern, the sensitivity of trajectory data 
necessitates that the privacy implications of their deployment 
are understood and assessed. The implementation of MIAs against 
four representative generative trajectory models establishes that 
such attacks are applicable in this domain, and that privacy risks 
are model-dependent and cannot be assumed absent simply because a 
model produces synthetic data. These findings highlight the need 
for systematic empirical privacy evaluation as a complement to 
formal guarantees, or as a substitute where such guarantees are 
absent.

As for future work, a broader range of adversarial privacy attacks 
and empirical evaluation methods should be explored to uncover 
additional privacy vulnerabilities in generative trajectory models. 
Such investigations will not only deepen the understanding of the 
privacy risks associated with these models but will also inform 
the development of more robust generative architectures and 
effective privacy-preserving defenses.

\bibliographystyle{IEEEtran}

\bibliography{ref}


\end{document}